\documentclass[11pt,a4paper]{article}
\usepackage{authblk}
\usepackage{dramatist}
\usepackage{multirow}
\usepackage{multicol}
\usepackage{subfigure}
\usepackage{acl2021}
\usepackage{latexsym}
\usepackage{graphicx}
\usepackage{subfigure}
\usepackage{hyperref}
\usepackage{mathptmx}
\usepackage{float}
\usepackage{url}
\usepackage{booktabs}
\usepackage{amsmath}
\usepackage{amsfonts}

\aclfinalcopy

\title{A Brief Study on the Effects of Training Generative Dialogue Models with a Semantic loss}

\author{Prasanna Parthasarathi\thanks{\ \ Corresponding author (pparth2@cs.mcgill.ca) \\ \ \ \ \  $^+$ Equal authorship}, \textsuperscript{$+$,1,4}
  Mohamed Abdelsalam, \textsuperscript{$+$,2,4}
  Joelle Pineau, \textsuperscript{1,3,5}
  Sarath Chandar \textsuperscript{3,4,5} \\
  \textsuperscript{1} School of Computer Science, McGill University
  \textsuperscript{2} University of Montr\'eal \\
  \textsuperscript{3} \'Ecole Polytechnique de Montr\'eal,\\
  \textsuperscript{4} Quebec Artificial Intelligence Institute (Mila),
  \textsuperscript{5} Canada CIFAR AI Chair
}

\begin{document}
\maketitle
\begin{abstract}
Neural models trained for next utterance generation in dialogue task learn to mimic the n-gram sequences in the training set with training objectives like negative log-likelihood (NLL) or cross-entropy. Such commonly used training objectives do not foster generating alternate responses to a context. But, the effects of minimizing an alternate training objective that fosters a model to generate alternate response and score it on semantic similarity has not been well studied. We hypothesize that a language generation model can improve on its diversity by learning to generate alternate text during training and minimizing a semantic loss as an auxiliary objective. We explore this idea on two different sized data sets on the task of next utterance generation in goal oriented dialogues. We make two observations (1) minimizing a semantic objective improved diversity in responses in the smaller data set (Frames) but only as-good-as minimizing the NLL in the larger data set (MultiWoZ) (2) large language model embeddings can be more useful as a semantic loss objective than as initialization for token embeddings. 
\end{abstract}

\section{Introduction}
\label{sec:intro}

Data for language generation tasks in goal-oriented dialogue has semantically diverse samples, where the diversity can be observed from the dialogue topics to the utterances used for getting information on specific slot-values from the user. But, in many niche domains, collecting a large high-quality annotated data set is costly, and often a small data set focused on specific tasks \cite{wei2018task,asri2017frames} is used for training. This restricts the model to only learn task-specific frequent contexts and seldom learn semantically similar context due to the lack of sufficient samples \cite{vinyals2015neural,serban2015survey,li2017adversarial,parthasarathi2018extending}. 

Optimizing only on objectives like negative log-likelihood (NLL), and Cross-Entropy (CE) losses foster learning by making the models mimic targets at the token level \cite{duvsek2020evaluating}. The models, hence, mostly generate only the observable patterns in the targets in training set \cite{huang2017parametric}. This can be attributed to the training procedure being uninformative about the semantic similarity of responses. To better understand, consider \texttt{Target: Would you like to travel to Paris ?, R1: How about Paris as your destination ?, R2: Would she like to read to me ? }. R2 has 4 tokens in the same position as in the target but R1 is semantically \textit{similar} to the target. However, the NLL/CE loss for predicting R2 will be lower than predicting R1. This is a common occurrence when training a language generation model, and training on a small data set can exacerbate this issue even further. 

Word embeddings from large language models like GloVe \cite{glove} , BERT \cite{devlin2018bert} or fastText \cite{fasttext} have been shown to have nice properties that preserve some of the linguistic structures \cite{sinha2020learning} that help in understanding semantic and temporal structures in dialogue. We make use of the semantics in the large word embeddings by computing a distance heuristic between the sampled text from model distribution and the target during training. This auxiliary semantic loss \footnote{\href{https://github.com/ppartha03/Semantic-Loss-Dialogue-Generation}{https://github.com/ppartha03/Semantic-Loss-Dialogue-Generation}} encourages the model in generating sentences that are similar to the target and thereby potentially diversifying the model responses. Although the results are on dialogue generation tasks, the results are comparable to any broad conditional language generation tasks like caption generation \cite{vinyals2015show}, text summarization \cite{luhn1958automatic} and others \cite{gatt2018survey}.

Our contributions in the paper are:
\begin{itemize}
    \item Comprehensively evaluate the proposed semantic loss on two differently sized data sets.
    \item Show that minimizing a semantic loss on the sampled responses as a training objective improves text generation diversity in limited data setting.
    \item Show that language model embeddings are useful as semantic loss than word embedding initialization.
\end{itemize}

\section{Conditional Language Generation}

In an encoder-decoder architecture, the encoder neural network \cite{rnnlm} encodes a textual summary of previous utterance exchanges between a user and an agent, $H_{i-1}$, and the current user utterance $u_i$. The encoded summary is used by a decoder network to generate the corresponding agent response ($a^*_i = (w_1^i, w_2^i, \ldots, w_T^i)$). 

Language generation models are mostly trained with NLL objective as defined in Equation \ref{nll},
\begin{equation}
    \mathbb{L}_{MLE} = -\sum_{t=1}^{T}\log P(w^i_t \mid w^i_{<t},H_{i-1},u_i)
    \label{nll}
\end{equation}
where $T$ is the number of tokens generated in the response ($a^*_i$), $w^i_t$ is the $t$-th word in the $i$-th utterance, and $w^i_{<t}$ denote tokens generated till step $t$.

\section{Semantic Loss}
\label{model}
We introduce training with a semantic loss computed with word embeddings from any trained language model. The semantic loss to be minimized is computed in three steps: (1) $a_{i}^{sampled} = (w_1^i, w_2^i, \ldots, w_{T'}^i)$ is generated by sampling tokens from decoder's distribution over the vocabulary at every step. (2) Average the word vectors of the sampled ($\hat{b}^{a^{sampled}_i}$) and ground truth responses ($\hat{b}^{a_i}$) with the embeddings from large language models like BERT, GloVe or fastText. Then, compute L2 distance between the two as shown in Equation \ref{dsem}. 
\begin{equation}
    d^i_{SEM} = \mid\mid \hat{b}^{a^{sampled}_i} - b^{a_i} \mid\mid_2
    \label{dsem}
\end{equation}
(3) Minimize $d^i_{SEM}$ calculated with the non-differentiable sampling operation, we use REINFORCE \cite{williams1992simple} to compute $\mathbb{L}_{SEM}$ (Equation \ref{semloss}).
\begin{equation}
    \mathbb{L}_{SEM} = - (-d^i_{SEM} - r(b)) \sum_{t=1}^{T'} \log P(w^{i}_t) 
    \label{semloss}
\end{equation}
where $T'$ is the number of tokens in $a_{i}^{sampled}$ and $r(b)$ is the reward baseline computed with average over a moving window of previous rewards to reduce the variance. The model minimizes $\mathbb{L}_{Train}$ as shown in Equation \ref{train_obj}.
\begin{equation}
    \mathbb{L}_{Train} =  \mathbb{L}_{MLE} + \alpha * \mathbb{L}_{SEM} 
    \label{train_obj}
\end{equation}
where $\alpha \in \mathbb{R}^+$ is a hyperparameter to specify the strength of the regularization by $\mathbb{L}_{SEM}$, the optimal value for $\alpha$ depends on the data set. 
Note: $\mathbb{L}_{Train}$ prefers $R1$ over $R2$ from the example in Section \ref{sec:intro}, unlike $\mathbb{L}_{MLE}$.

\section{Experiments}

We experiment on two differently sized data sets -- Frames \cite{asri2017frames} and MultiWoZ 2.0 \cite{budzianowski2018multiwoz} -- which are relatively small and large. We compute $\mathbb{L}_{SEM}$ using the commonly used language model embeddings BERT-Base \cite{devlin2018bert}, GloVe \cite{glove} and fastText \cite{fasttext} to compare the benefit of using different embeddings.

\textbf{Evaluation Metrics:} We measure the performance on overlap based metric BLEU \cite{papineni2002bleu}; and diversity in the generated text by computing the fraction of distinct-1 and distinct-2 grams, similar to \citet{welleck2019neural,li2015diversity}, on validation set. Also, as a proxy to evaluate generalization to generating n-grams that the decoder was never trained to, we measure the fraction of bigrams generated by the model during validation that were not in the training targets, as \textit{\% Unseen}. Also, to measure the effects of minimizing the semantic loss on language quality, we perform human evaluation for comparing the different training techniques. Further we compare the improvements in diversity between using BERT for initialization of word embeddings and using it in a semantic loss objective.

\subsection{Quantitative Evaluation}

\begin{figure*}[h!t]
\centering
\subfigure[BLEU on Frames]{
    \includegraphics[width=0.65\columnwidth]{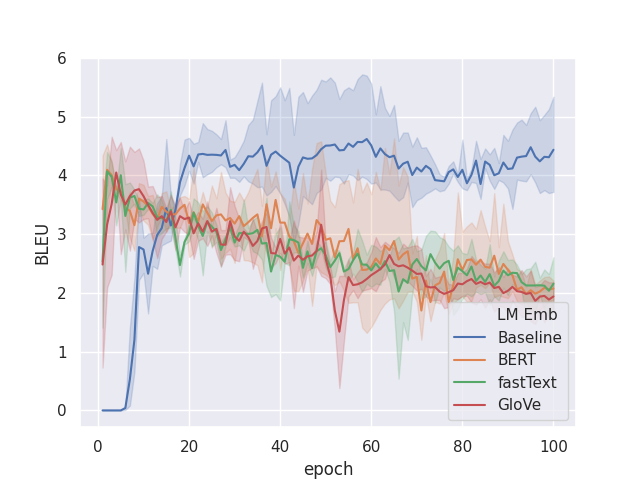}
    \label{sub_bleu_frames}
}
%
%
\subfigure[Distinct-2 on Frames]{
    \includegraphics[width=0.65\columnwidth]{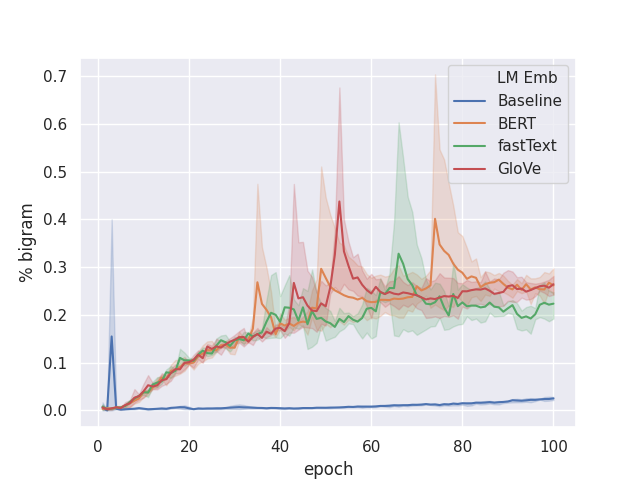}
    \label{sub_distinct_2_frames}
}
\subfigure[Unseen Bigrams on Frames]{
    \includegraphics[width=0.65\columnwidth]{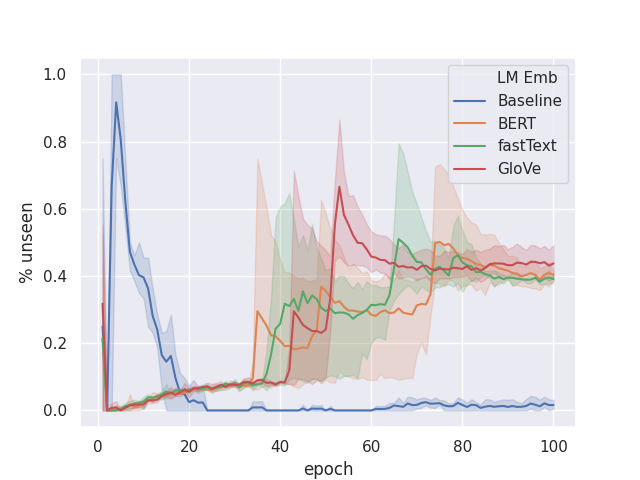}
    \label{sub_unseen_bi_frames}
}
%

\subfigure[Distinct-2 BERT Initialization and finetuning vs BERT in semantic loss on Frames]{
    \includegraphics[width=0.65\columnwidth]{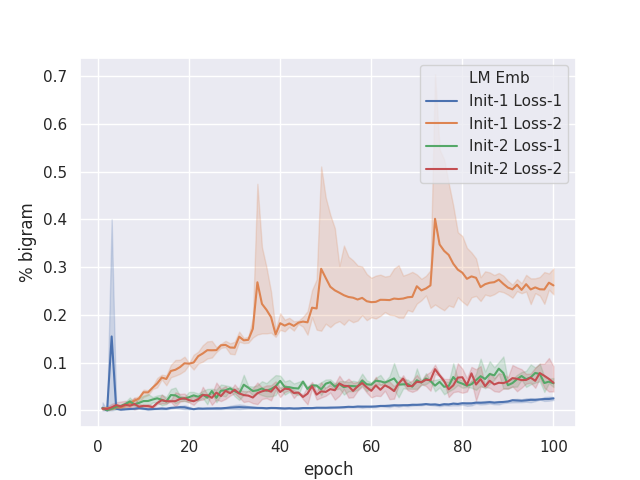}
    \label{sub_bert_init_frames}
}
\subfigure[BLEU on MultiWoZ]{
    \includegraphics[width=0.65\columnwidth]{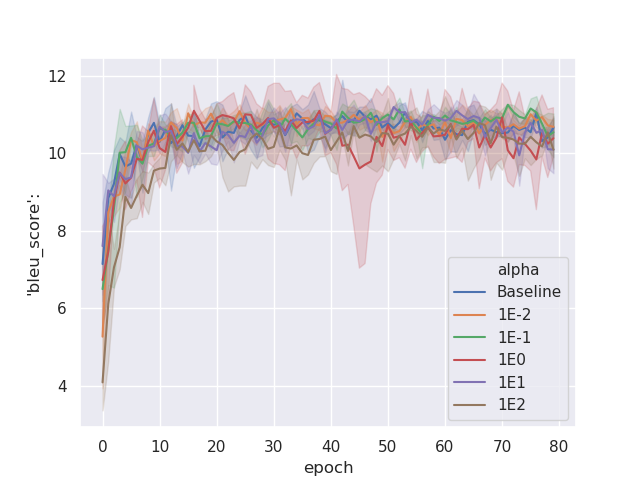}
    \label{sub_bleu_mwoz}
}
\subfigure[Distinct-2 on MultiWoZ]{
    \includegraphics[width=0.65\columnwidth]{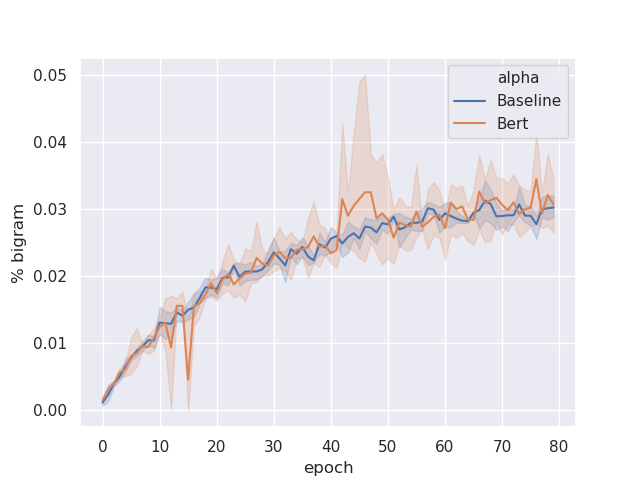}
    \label{sub_distinct_2_mwoz}
}
\caption{Quantitative comparison of different loss functions, and initialization on Frames and MultiWoZ data sets.}
\label{none}
\end{figure*}

Experimental result in Figure \ref{sub_bleu_frames} shows that performance of the model trained with $\mathbb{L}_{Train}$ decreases on the overlap based metric, BLEU. This is explained by the $\mathbb{L}_{Train}$ trained models, with greedy decoding, generating a greater fraction of unique bigrams (Figure \ref{sub_distinct_2_frames}) on the validation set than the $\mathbb{L}_{MLE}$ trained model: measured with metrics distinct-1 and distinct-2 \cite{li2015diversity}. As the model learns to discover semantically similar bigrams, the performance on overlap based metric decreases.
Further, \% Unseen metric measured in Figure \ref{sub_unseen_bi_frames} shows that $\mathbb{L}_{Train}$ fosters generation of new bigrams.


In the experiments, we observed a pattern of \% Unseen spiking at regular intervals, indicating that the loss helped the model to periodically discover newer bigrams, which increased the NLL in training as the syntax around the bigram has to be relearned by minimizing the now higher NLL objective.
\begin{table}[h!t]
    \centering
    \small
\begin{drama}
  \Character{User}{user}
  \Character{Semantic Beam1}{beama}
  \Character{Semantic Beam2}{beamb}
  \Character{Semantic Beam3}{beamc}
  \Character{Semantic Beam4}{beamd}
  \Character{Semantic Beam5}{beame}
  \Character{Target}{target}
  
  \userspeaks: of those 3 options , i would prefer 11 days any other hotel options can you check if there are other hotel options for september 1 - 20 ? what are the departure and return dates for this.  
  \targetspeaks: sept 13th through the 19th.
  \beamaspeaks: i ’ m sorry i have nothing from santiago .
  \beambspeaks: i ’ m sorry i have nothing from santiago . is there another destination and would you be interested
  \beamcspeaks: i ’ m sorry i have nothing from santiago . is there another destination ?
  \beamdspeaks: i ’ m sorry i have nothing from santiago . is there another destination you would like to go
  \beamespeaks: i ’ m sorry i have nothing from santiago . is there another destination you would like to be
\end{drama} 
\par\noindent\rule{0.45\textwidth}{0.4pt}
         \begin{drama}
  \Character{User}{user}
  \Character{Beam 1}{beama}
  \Character{Beam 2}{beamb}
  \Character{Beam 3}{beamc}
  \Character{Beam 4}{beamd}
  \Character{Beam 5}{beame}
  \Character{Target}{target}

  \userspeaks: of those 3 options , i would prefer 11 days any other hotel options can you check if there are other hotel options for september 1 - 20 ? what are the departure and return dates for this. 

  \targetspeaks: sept 13th through the 19th.
  \beamaspeaks: i can i do not have to help , sorry , i sorry , sorry , i sorry ,
  \beambspeaks: i can i do n't have to help , sorry , i sorry , i sorry , i sorry
  \beamcspeaks:  i can i do not have to help , sorry , i sorry , i sorry , i sorry
  \beamdspeaks: i can i do not have for that sorry , i sorry , i sorry , i sorry ,
  \beamespeaks: i can i do not have to help , sorry , i sorry , i sorry , sorry ,
\end{drama}
\caption{Comparing the diversity in beam search between the model trained with $\mathbb{L}_{Train}$ (top) and with $\mathbb{L}_{MLE}$ (bottom)}
    \label{tab:beam_comparison_frames}
\end{table}
This is different from beam search as beam sampling conforms to the distribution learnt with $\mathbb{L}_{MLE}$, whereas $\mathbb{L}_{Train}$ allows to learn a distribution that allows learning to use valid alternatives in the training. This allows a better beam search, as shown in the example Table \ref{tab:beam_comparison_frames}.

\subsection{BERT Initialization vs BERT Semantic loss}
We construct 4 different models by combining the two different loss functions (Loss1: $\mathbb{L}_{MLE}$, Loss2: $\mathbb{L}_{Train}$) with two different initializations (Init1: random, and Init2: BERT) for the word embeddings. Diversity measured with {\it distinct-2} (Figure \ref{sub_bert_init_frames}) showed that \textit{Init1;Loss2} model showed greater improvements than \textit{Init2;Loss1} or \textit{Init2;Loss1}. The result suggests that BERT can be more useful in $\mathbb{L}_{Train}$ than embedding initialization. This could be reasoned by the strong regularization enforced by the word embedding that is unyielding to exploration in generating sequences in addition to the $\mathbb{L}_{MLE}$ objective.


\subsection{Negative Result in MultiWoZ}
We observed that the model trained with $\mathbb{L}_{Train}$ performed only as good as training with $\mathbb{L}_{MLE}$ on our defined evaluation metrics (Figure \ref{sub_bleu_mwoz},\ref{sub_distinct_2_mwoz}) in MultiWoZ. The overlap based metric and unique bigrams generated did not have as much improvement as it had in Frames data set (Figures \ref{sub_distinct_2_frames}, \ref{sub_distinct_2_mwoz}).
\begin{table}
\small
 \begin{drama}
  \Character{User}{user}
  \Character{Best BLEU}{agentb}
  \Character{Diverged}{agenta}
  \userspeaks: i will also need a taxi to pick me up by 24:30 . i need the contact number and car type please.  
  \agentbspeaks: i have booked you a yellow lexus . the contact number is 07346991147.
  \agentaspeaks: okay pull d assisting joining botanic gardens , good and good bye.
\end{drama}
\caption{Aggressively exploring with dropping larger fraction of tokens in a sentence lead to divergence in language generation in MultiWoZ as shown.}
\label{tab:mwoz-negative-main}
\end{table}

To overcome this issue, during training, we increased the model's exploration to newer tokens by masking tokens in the decoder output at random before sampling a response. This helped the model in discovering newer bigrams eventually. This technique generated larger fraction of unseen bigrams but the randomness in dropping tokens generated more noise in the text generated (Table \ref{tab:mwoz-negative-main}). Making the random exploration useful with additional constraints to keep the syntax from diverging is a potential future work.
\subsection{Human Evaluation}
We perform two human studies (Appendix \ref{human_study}) with two sets of 100 randomly sampled contexts from test set of Frames data set with 3 scorers per pair.
\begin{table}[!h]
\centering
\footnotesize
\begin{tabular}{cccc}
\toprule
  Metric & \% Wins & \% Losses&  \% Ties \\
\midrule
{\it Diversity} &{\bf 65}& 16 & 19\\
\midrule
{\it Relevance} &{\bf 45}& 38& 17\\
\bottomrule
\end{tabular}
\caption{\textbf{Study 1:} \%Wins denote the \#times the scorers picked \emph{Init1;Loss2}'s response and \%Loss is when it was the \emph{Init1;Loss1}'s response.}
\label{tab:human_eval_study1}
\end{table}
\begin{table}
\centering
\footnotesize
\begin{tabular}{cccc}
\toprule
  Metric & \% Wins & \% Losses&  \% Ties \\
\midrule
{\it Diversity} &\textbf{63}& 24 & 13\\
\midrule
{\it Relevance} &\textbf{41}& 31& 28\\
\bottomrule
\end{tabular}
\caption{\textbf{Study 2:} \%Wins denote the \#times the scorers picked \emph{Init1;Loss2}'s response and \%Loss is when scorers picked the \emph{Init2;Loss1}.}
\label{tab:human_eval_study2}
\end{table}
In Study 1, the volunteers were shown the responses generated with \emph{Init1;Loss1} and \emph{Init1;Loss2}. Like in \citep{li2015diversity}, we ask the volunteers to select the one that is \textit{relevant} to the context, and the one that is \textit{interesting}/\textit{diverse} in two separate questions. We allow ties in both the questions. In Study 2, we compare \emph{Init2;Loss1} and \emph{Init1;Loss2} with questions as in Study 1.

The results of Study 1 and Study 2 shown in Table \ref{tab:human_eval_study1} and \ref{tab:human_eval_study2} show that, despite the lower BLEU scores, minimizing $\mathbb{L}_{Train}$ indirectly fosters diversity in responses; human scorers found the model trained with the proposed semantic loss objective to be diverse/interesting on an average of {\it 65\%} and {\it 63\%} in studies 1 and 2 respectively. This verifies again in a different experiment that BLEU scores do not correlate well with human scores \cite{liu2016not}. The regularization from the BERT initialization is not promoting diversity which, from the experiments, depends on minimizing the semantic objective. The relevance of the response is not significantly higher than the baseline, which was expected as the semantic loss was expected only to improve the diversity. 

\section{Conclusion}
\label{conclusion}
Training with a semantic loss has a positive effect in a smaller data set and that reflects on the model's improvement in diversity measuring metrics. But, the semantic loss was not very effective in a large data set due to the lack of diversity within and a hard bias dictated by the samples in the data set. The results obtained in the paper shows that training with semantic loss can be effective in low data setting. 

\section*{Acknowledgements}
We would like to acknowledge Compute Canada and Calcul Quebec for providing computing resources used in this work. The authors would also like to thank members of Chandar Research Lab, Mila for helping with the code reviews and reviewing the manuscripts. Sarath Chandar and Joelle Pineau are supported by Canada CIFAR AI Chair, and Sarath Chandar is also supported by an NSERC Discovery Grant.

\bibliography{refs}
\bibliographystyle{acl_natbib}

\appendix

\section{Training and hyperparameters}
\begin{itemize}
    \item We used a 128 unit hidden size LSTM with a 128 unit input embedding dimension.
    \item The range of the $\alpha$ we tested in log-scale is [-2,2]. And, the best alpha selected based on the early saturation of distinct-2 was 1E-1 and used this for experiments in different language model embeddings used for computing $\mathbb{L}_{SEM}$.
    \item We use Adam optimizer with 4E-3 as learning rate and other parameters as default. 
    \item For the choice of word embeddings, we used 300 dimensional GloVe and fastText, and 768 dimensional BERT-Base. 
    \item For REINFORCE with baseline, we computed the average for the last 20 samples as the baseline.
    \item We averaged the results over 5 different seeds. For the baseline model, we chose the best performing seed with respect to BLEU score and for the model trained with $\mathbb{L}_{Train}$ based on early saturation on distinct-2 on the validation set for human evaluation.
\end{itemize}

\section{Frames Experiments}

\subsection{Word repeats}
\begin{figure}[h!t]
    \centering
    \includegraphics[width=0.8\columnwidth]{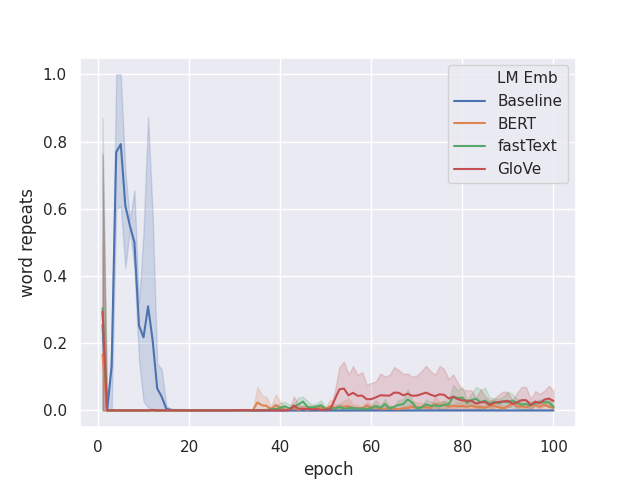}
    \caption{Comparing $d_{SEM}$ with different word embeddings on fraction of bigrams generated on the validation set that are word repeats on the Frames data set.}
    \label{fig:frames_word_repeats}
\end{figure}

Evaluating generalization to unseen bigrams is tricky as there can be potentially many word repeats. To not count that, we looked at the fraction of bigrams that were word repeats, one of the most common errors by language generation models (Figure \ref{fig:frames_word_repeats}). 

The result showed two interesting things: First, the word repeats are minimal but does happen when training with semantic loss, though the gain of discovering unseen bigrams is more useful. Second, the NLL trained model initially generates many word repeats along with a few unseen tokens and they both die down due to the strong MLE objective that overfits to the targets in the training.

\subsection{Human Evaluation}
\label{human_study}

For human evaluation, we asked for English speaking graduate students as volunteers to take part in the two studies. To reduce the cognitive load on individual participants, we split the 100 samples in 4 sets of 25 samples. We computed the inter-annotators agreement with cohen-kappa coefficient \cite{cohen1960coefficient} in the sklearn package \cite{scikit-learn}.

\begin{table}[H]
    \centering
    \begin{tabular}{ccc}
    \toprule
         & \textbf{Q1:Relevance} & \textbf{Q2:Diversity}\\
    \toprule
         \textit{Study 1} &0.28  &0.22\\
    \midrule
         \textit{Study 2} & 0.33 & 0.23\\
    \bottomrule
    \end{tabular}
    \caption{Average of cohen kappa score averaged over the evaluation of annotators on the different sets of samples in the two studies.}
    \label{tab:iaa_study}
\end{table}

The results shown in Table \ref{tab:iaa_study} that the annotators had a fair agreement in the two studies. The range of the scores is between -1 and 1, and a score above 0 indicates agreement amongst the annotators. The slightly lower agreement on Q2 is because of the ambiguity in the perception of "what is interesting".

\section{MultiWoZ Experiments}
\label{app:multiwoz_exp}

\subsection{Negative Result}
\label{app:mwoz_negative}
We observed that the semantic loss was not as useful as it was in the smaller data set. The bigram distribution of the two data sets (Table \ref{tab:data set_context} and \ref{tab:data set_target}) showed that the bigrams in the context on an average occurs 92 times in  MultiWoZ as compared to only 17 times in Frames. Similarly, a bigram in the target occurs 13 times in MultiWoZ compared to only 5.4 times in Frames. 

From the analysis on the distribution of bigrams in the two data sets, we arrived at the following conjecture: With a simplistic assumption, consider the following sentences: \texttt{I want to leave \textbf{from London}}, \texttt{I want to leave \textbf{on Tuesday}}, \texttt{I want to leave from Florida} occur 3, 2, and 5 times respectively in a small data set and 30, 20, and 50 times in a relatively larger data set. The language model of the decoder, after generating \texttt{I want to leave}, will sample one of the three bigrams, \texttt{on Tuesday, to London, from Florida}.

\begin{table}[h!t]
    \centering
    \begin{tabular}{ccc}
    \toprule
      data set & Unique Bigrams & Total Bigrams \\
    \midrule
    {\it Frames} & ~30K & ~0.5M \\
    \midrule
    {\it MultiWoZ} & ~40K & ~3.6M \\
    \bottomrule
    \end{tabular}
    \caption{Count of Bigrams from only the contexts of the two data sets}
    \label{tab:data set_context}
\end{table}

\begin{table}[h!t]
    \centering
    \begin{tabular}{ccc}
    \toprule
      data set & Unique Bigrams & Total Bigrams \\
    \midrule
    {\it Frames} & ~22K & ~127k \\
    \midrule
    {\it MultiWoZ} & ~71K & ~900k \\
    \bottomrule
    \end{tabular}
    \caption{Count of Bigrams from only the targets of the two data sets}
    \label{tab:data set_target}
\end{table}

The output of the encoder-decoder at every step being a mulitnomial distribution over the vocabulary, the architecture can be abstracted for our understanding to maintain a Dirichlet distribution that is generalizable.

The bias of sampling \texttt{from Florida} is much higher in a large data set and relatively much lower in a smaller data set, which can even generate \texttt{I want to leave from Florida to London on Tuesday} with a relatively higher probability. As sampling from the decoder is still dependent on $\mathbb{L}_{MLE}$, the diversity in sampling is decreased when training with NLL on a large data set.

\begin{figure}[h]
    \centering
    \includegraphics[width=0.8\columnwidth]{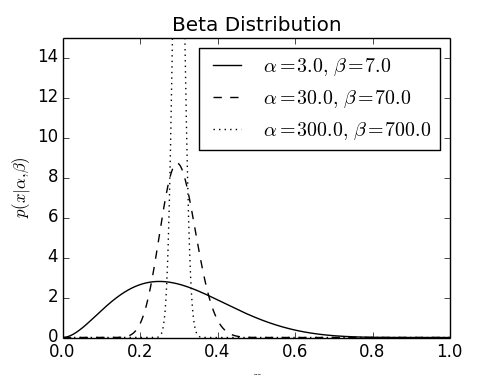}
    \caption{Beta distribution with differently scaled $\alpha$ and $\beta$ values. The lower values correspond to the smaller data sets and the higher values correspond to the larger data sets.}
    \label{fig:beta}
\end{figure}

But then, as the larger data set has 7 times more support for a bigram than in the smaller data set, out of distribution sampling is difficult.

\subsection{Out-of-NLL Sampling}

To break the rigid sampling distribution, with a non-zero probability we dropped words from the vocabulary before sampling the tokens in $a_i^{sampled}$. 

\begin{figure}[h!t]
    \centering
    \includegraphics[width=0.8\columnwidth]{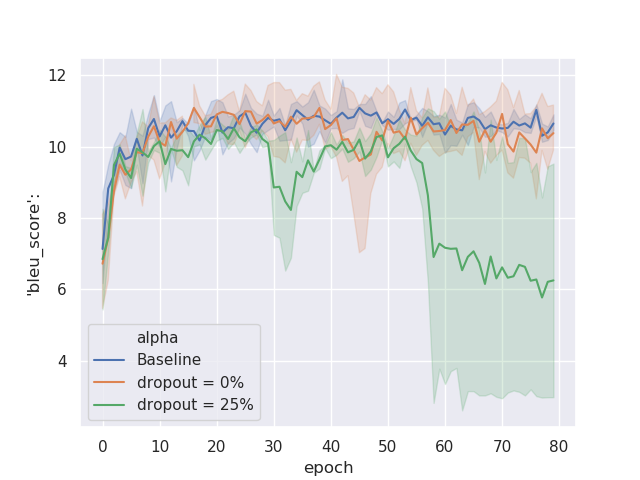}
    \caption{Effect of dropout on automatic evaluation metric. The drop in BLEU is due to the model generating newer bigrams.}
    \label{fig:mwoz_dropout_bleu}
\end{figure}

\begin{figure}[H]
    \centering
    \includegraphics[width=0.8\columnwidth]{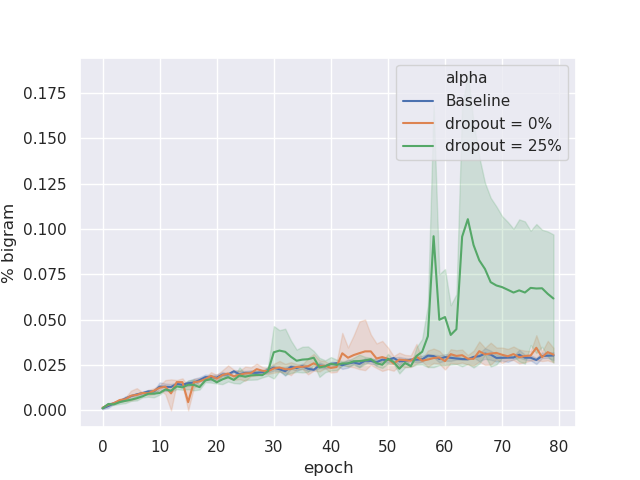}
    \caption{The distinct-2 grams when trained with dropout and random substitution indeed helped the model to sample out-of-NLL distribution. But, the overwhelming noise diverged the training and the model responses degenerated.}
    \label{fig:mwoz_dropout_distinct2}
\end{figure}

With the semantic loss providing non-binary scores, the model gets feedback for all sampled responses, even those unlikely to be sampled but are sampled due to the masking of the vocabulary. That lead to a sharp divergence of training (Table \ref{tab:mwoz-negative-main}) even before the model learnt to appropriately diversify its responses (Figure \ref{fig:mwoz_dropout_distinct2}).

The {\it \% unseen} and distinct-1 and 2 scores keep increasing (Figures \ref{fig:mwoz_dropout_distinct2}) but due to the high amount of diversity in the tokens generated, many of the responses were not legible as seen in Table \ref{tab:mwoz-negative-main}.

\end{document}